\documentclass[sigconf]{acmart}
\usepackage{multirow}
\usepackage{multicol}
\usepackage{graphicx}
\usepackage{subfigure} 
\usepackage{bbding}
\usepackage{makecell}
\usepackage{fancyhdr}

\renewcommand\footnotetextcopyrightpermission[1]{} 
\AtBeginDocument{%
  }

\begin{document}

\title{T-Stars-Poster: A Framework for Product-Centric Advertising Image Design}

\author{Hongyu Chen}
\authornote{Both authors contributed equally to this research.}
\email{yinchen.chy@alibaba-inc.com}
\affiliation{%
  \institution{Alibaba Group}
  \city{Beijing}
  \country{China}
}

\author{Min Zhou}
\authornotemark[1]
\email{yunqi.zm@alibaba-inc.com}
\affiliation{%
  \institution{Alibaba Group}
  \city{Beijing}
  \country{China}
}

\author{Jing Jiang}
\authornotemark[1]
\authornote{Work done during the internship at Alibaba Group.}
\email{jiangjing1998@bupt.edu.cn}
\affiliation{%
  \institution{Beijing University of Posts and Telecommunications}
  \city{Beijing}
  \country{China}}

\author{Jiale Chen}
\email{cjl414939@taobao.com}
\affiliation{%
  \institution{Alibaba Group}
  \city{Beijing}
  \country{China}
}

\author{Yang Lu}
\email{ly430273@alibaba-inc.com}
\affiliation{%
  \institution{Alibaba Group}
  \city{Beijing}
  \country{China}
}

\author{Bo Xiao}
\email{linuxnms@gmail.com}
\affiliation{%
  \city{Beijing}
  \country{China}
}

\author{Tiezheng Ge}
\authornote{Corresponding author.}
\email{tiezheng.gtz@alibaba-inc.com}
\affiliation{%
  \institution{Alibaba Group}
  \city{Beijing}
  \country{China}
}

\author{Bo Zheng}
\email{bozheng@alibaba-inc.com}
\affiliation{%
  \institution{Alibaba Group}
  \city{Beijing}
  \country{China}
}


\begin{abstract}
Creating advertising images is often a labor-intensive and time-consuming process. Can we automatically generate such images using basic product information like a product foreground image, taglines, and a target size? Existing methods mainly focus on parts of the problem and lack a comprehensive solution. To bridge this gap, we propose a novel product-centric framework for advertising image design called T-Stars-Poster. It consists of four sequential stages to highlight product foregrounds and taglines while achieving overall image aesthetics: prompt generation, layout generation, background image generation, and graphics rendering. Different expert models are designed and trained for the first three stages: First, a visual language model (VLM) generates background prompts that match the products. Next, a VLM-based layout generation model arranges the placement of product foregrounds, graphic elements~(taglines and decorative underlays), and various nongraphic elements~(objects from the background prompt). Following this, an SDXL-based model can simultaneously accept prompts, layouts, and foreground controls to generate images. To support T-Stars-Poster, we create two corresponding datasets\footnote{The primary dataset is available at \href{https://github.com/alimama-creative/PITA-dataset}{https://github.com/alimama-creative/PITA-dataset}.} with over 50,000 labeled images.  Extensive experiments and online A/B tests demonstrate that T-Stars-Poster can produce more visually appealing advertising images.
\end{abstract}

\begin{teaserfigure}
  \centering
  \includegraphics[width=0.95\linewidth]{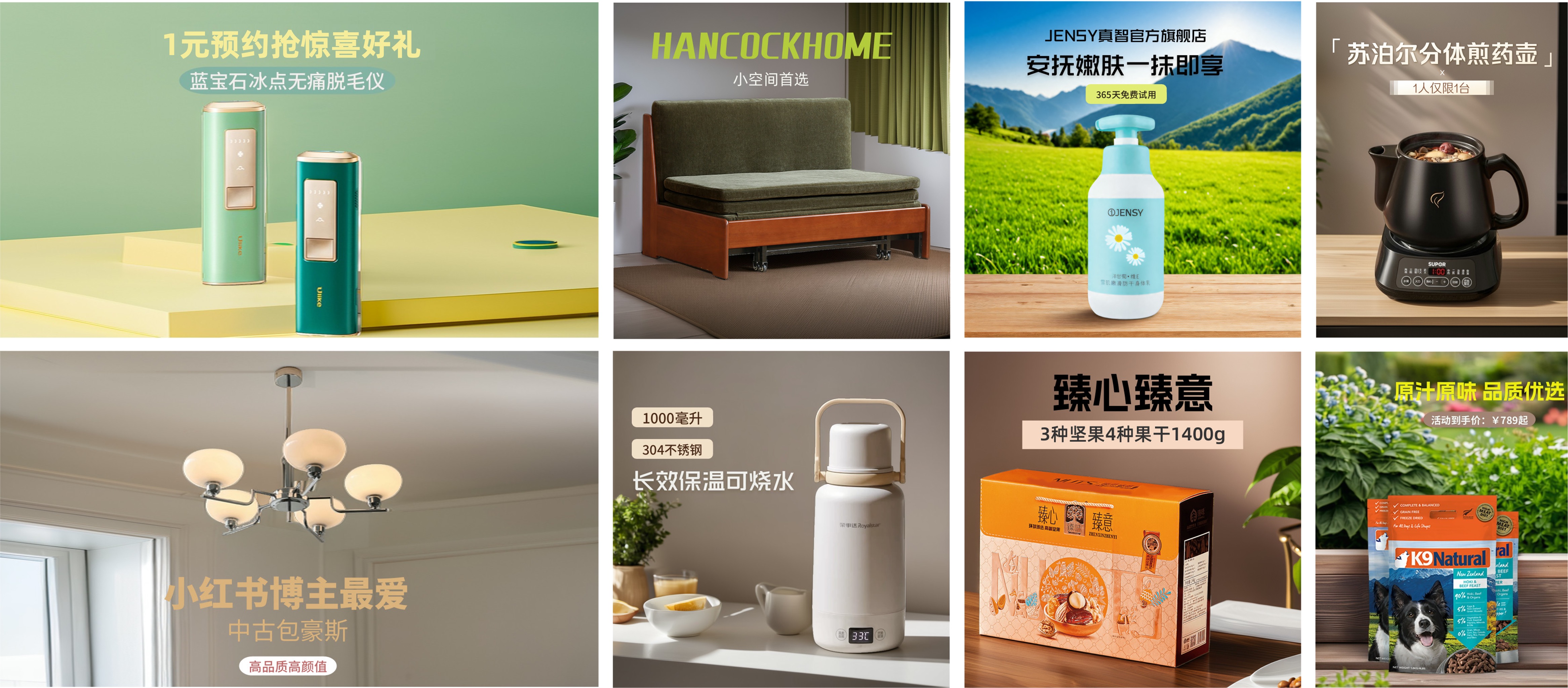}
  \caption{Generated advertising images by our methods with a product foreground image, taglines, and a target size as input.}
  \label{fig:teaser}
\end{teaserfigure}

\maketitle

\section{Introduction}
Advertising images~(as shown in Figure~\ref{fig:teaser}) are essential for commercial recommendation. Visually appealing images usually have high click-through rates~\cite{wang2021hybrid} but are labor-intensive and time-consuming to create~\cite{DBLP:conf/kdd/VashishthaPMNAK24}. With the rapid advancement of generative methods~\cite{menick2018generating,van2016pixel,ho2020denoising,Rombach2021HighResolutionIS}, it is now possible to automatically generate advertising images using only basic product inputs: a product foreground image, marketing taglines, and a target size.

Currently, generating advertising images using only the basic inputs has rarely been explored. Existing methods~\cite{du2024towards,weng2024desigen,wang2024prompt2poster,lin2023autoposter,zhou2022composition} mainly focus on parts of the problem, such as background or layout generation, rather than offering a complete solution for integrating product foregrounds and taglines. Although these methods can be combined, they struggle to effectively highlight product images and taglines while maintaining visual appeal and harmony. Specifically, building on existing approaches~\cite{du2024towards,weng2024desigen,wang2024prompt2poster,lin2023autoposter,zhou2022composition}, we can first utilize predefined rules to create background prompts, which are then fed into text-to-image inpainting models to complete backgrounds for product foregrounds. Layouts for taglines and decorative elements~(collectively referred to as graphic elements) are predicted based on the images and taglines. Finally, attribute prediction and graphics rendering techniques are employed to add these graphic elements to the images. However, this process has several limitations: a) Lack of adaptability: The background prompts cannot match the specific content and shape of the product, leading to poor foreground-background harmony. b) Fixed foreground positioning: The foreground position cannot be adjusted based on the product characteristics and the background prompt, potentially resulting in inappropriate product size and poor composition. c) Restricted tagline placement: The whole image content is set before taglines are added, limiting space for tagline placement.

\begin{figure*}
    \centering
    \vspace{-0.2cm}
    \includegraphics[width=0.93\linewidth]{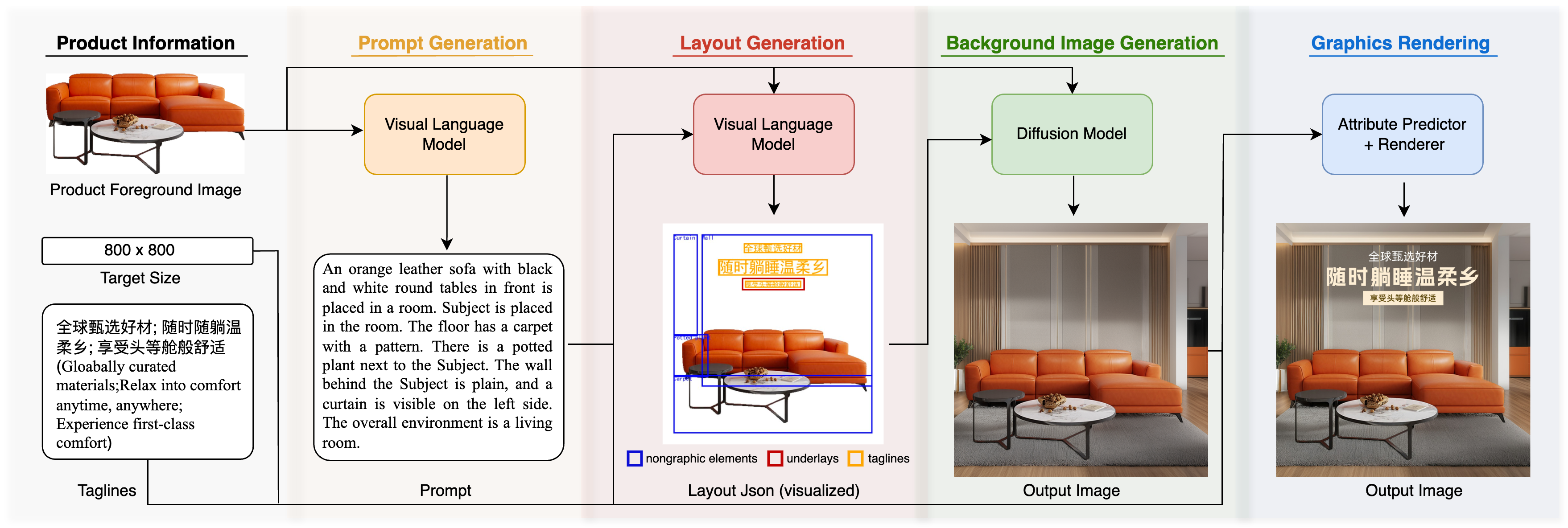}
    \vspace{-0.2cm}
    \caption{Pipeline of T-Stars-Poster. It consists of four stages and generates advertising images centered around product information.}
    \vspace{-0.2cm}
    \label{fig:pipeline}
\end{figure*}

In this paper, we propose a novel product-centric framework called T-Stars-Poster for advertising image generation using only basic product inputs. We refine the stages and task definitions to enhance foreground-background compatibility and the spatial arrangement of product foregrounds and taglines.

As illustrated in Figure~\ref{fig:pipeline}, we first introduce a task to generate prompts, using a Visual Language Model (VLM)~\cite{dong2024internlm} to create background prompts that match the scene, placement, and angle with product foregrounds. Next, a layout generation task arranges the graphic elements (marketing taglines and decorative underlays), product foreground, and other nongraphic elements~(objects to be generated from the background prompt). 
This arrangement is based on the product foreground, background prompt, taglines, and target size. 
By arranging the product foreground and taglines simultaneously, rather than generating the whole image before deciding the graphic layout, we reduce layout constraints and present product information more effectively. Furthermore, considering nongraphic elements together helps minimize conflicts with taglines that could affect readability and attractiveness. We then train an inpainting text-to-image model based on SDXL~\cite{podell2023sdxl} that can be guided by input layouts. Combining inpainting, layout, and prompt control effectively is challenging due to potential conflicts among multiple control signals. A LoRA~\cite{hu2021lora} adaption strategy is adopted to balance conflicts and facilitate training. Since layout mainly influences higher-level abstract information, we apply layout control only in the deeper blocks of the UNet~\cite{DBLP:conf/cvpr/HeZRS16}, along with a multi-scale training strategy to enhance stability and efficiency. Finally, we use an attribute prediction and graphics rendering module to overlay graphic elements onto the image.
To build the framework and verify its effectiveness, we have collected and labeled over 50,000 images for datasets. In test, T-Stars-Poster shows it can produce advertising images with better visual quality. It has been deployed on two advertising recommendation scenarios of Taobao, and online A/B tests show that its generated images are more appealing to users.

In summary, our main contributions are as follows:
\begin{itemize}
    \item We propose a framework T-Stars-Poster for advertising image design given only a product foreground, taglines, and a target size. It centers on product foregrounds and taglines to improve the product prominence and image aesthetics.
    \item A VLM is introduced to generate suitable prompts for image generation using the input product foreground image, considering product placement angle and shape.
    \item To arrange product foregrounds and taglines more effectively, a VLM-based layout generator is proposed. It can mix and arrange graphic and nongraphic elements with the target size and prompt. Accordingly, an image generation model that simultaneously accepts layout, prompt, and foreground control is proposed. Several strategies are introduced to enhance effectiveness and efficiency.
    \item We have collected and labeled over 50,000 images to create datasets for advertising image generation. The effectiveness of our method has been validated through tests on this dataset and online A/B tests in real advertising scenarios.

\end{itemize}

\section{Related Work}
\subsection{Advertising Image Design}
Many methods have been proposed to automate advertising image design~\cite{zhou2022composition,hsu2023posterlayout,li2023relation,seol2024posterllama,lin2023autoposter,li2023planning}, falling into two categories: background-based and foreground-based. Background-based methods~\cite{zhou2022composition,hsu2023posterlayout,li2023relation,seol2024posterllama,lin2023autoposter} predict graphic layouts for complete images~(not product foregrounds), followed by attribute prediction and graphics rendering to create the final image. These methods face spatial constraints due to fixed image content, limiting tagline content. In contrast, P\&R~\cite{li2023planning} is a foreground-based method using both foreground images and tagline content to predict layout, then outpainting the background and rendering taglines according to the given background prompt and generated layout. The tagline content and layout are more flexible. However, P\&R ignores the harmony between the background prompt and the generated layout. The layout may conflict with background generation. Our framework generates the overall layout for graphic elements, product foreground, and other elements jointly, using product foregrounds, background prompts, and tagline content to ensure harmony.

\subsection{Controllable Image Generation}
There have been significant developments in text-to-image models~\cite{Rombach2021HighResolutionIS,podell2023sdxl}. To control instances and improve image quality, some methods introduce layout guidance, dividing into training-free and training-based categories. Training-free methods adjust the attention map or compute attention-based loss during inference~\cite{kim2023dense, 
chefer2023attend, xie2023boxdiff, chen2024training, bar2023multidiffusion}. Training-based methods~\cite{li2023gligen, wang2024instancediffusion} add gated self-attention layers to models and train them with corresponding data.

For advertising image generation~\cite{du2024towards, wang2023generate, cao2024product2img, li2023planning}, text-to-image models with inpainting capabilities are often utilized. Inpainting preserves foregrounds and prompts describe the backgrounds. 
The foreground image is integrated using extra channels~\cite{Rombach2021HighResolutionIS} or an inpainting ControlNet~\cite{zhang2023adding}. 
To better manage the layout and enhance image quality, we combine inpainting with layout control. Currently, there is limited exploration in this area. SceneBooth~\cite{scenebooth} is the first to integrate both controls in training, combining gated self-attention layers and inpainting ControlNet to train a model based on SD 1.5~\cite{Rombach2021HighResolutionIS}. However, directly combining the two to train a larger model like SDXL can lead to non-convergence or a decline in the quality of generated images. Therefore, we use several strategies to ensure training stability, generation quality, and efficiency.

\section{PITA Dataset and PIL Dataset}
\label{dataset}
We collect a Product-Centric Image-Tagline Advertising (PITA) dataset with 38,017 samples from Taobao
and the CGL dataset~\cite{zhou2022composition}. 1,000 of them are for test. PITA covers major categories of Taobao, such as clothing, food, furniture, and electronics.
It features images with four distinct aspect ratios: 0.684, 1.0, 0.667, and 0.75. In the collection process, we exclude images with messy backgrounds, plain colors without shadows, unattractive stickers, close-ups, poorly presented taglines, or cluttered elements. Each has labels for prompts (foreground and background captions), a product mask, and a layout. Each element is represented with a type and a bounding box (bbox). Graphic elements contain ``Logo'', ``Tagline'', and ``Underlay''. Initial annotations are generated using automatic methods such as DAMO Academy's matting API~\cite{damo_matting} for foreground extraction, GPT-4o~\cite{gpt4o} for image captioning, a detection model~\cite{DBLP:conf/cvpr/DaiCX0LY021} trained with the CGL dataset and Grounding-DINO~\cite{liu2024grounding} for element detection, and an OCR model~\cite{ppocrv4} for tagline extraction. Annotations are then manually reviewed and corrected for accuracy.

We also develop a Product-Centric Image Layout (PIL) dataset for layout-controlled background inpainting. It contains images without graphic elements and includes 12,247 samples, with 1,000 reserved for testing. The filtering and annotation process is the same as the PITA dataset, excluding graphic elements.

\section{Method}
\subsection{Overall Framework}
As illustrated in Figure~\ref{fig:pipeline}, our proposed T-Stars-Poster framework consists of four stages: prompt generation, layout generation, background image generation, and graphics rendering.

\subsection{Prompt Generation}
For prompt generation, we fine-tune a large vision-language model $\pi_{prompt}$  based on InternLM-XComposer2-vl (XCP2)~\cite{dong2024internlm}. The method is illustrated in the upper part of Figure~\ref{fig:layoutGen}. 

\textbf{Input data format.} We describe the prompt generation task and request foreground/background descriptions. Meanwhile, we feed the product foreground into the visual encoder.

\textbf{Output data format.} To support layout and background image generation tasks and understand the foreground-background relationship, the prompt generation model $\pi$ predicts both the foreground description $p_{fore}$ and background description $p_{back}$ simultaneously. We format the output in JSON as it is compatible with the pre-trained VLM and simplifies the following analysis. The output JSON is structured as $p = (p_{fore}, p_{back})$.

\textbf{Training scheme.}
We use a VLM trained after supervised fine-tuning~(SFT)~\cite{liu2024visual}, allowing the model to understand images and follow instructions. We only fine-tune the LLM part with our dataset for the prompt generation task, using cross-entropy loss. The model $\pi_{prompt}$ takes a foreground image $I_{fore}$ and predefined instructions to generate a foreground description and predict a background description. The prompt generation process is as follows:

\begin{align}
p  = (p_{fore}, p_{back}) = \pi_{prompt}(I_{fore}) .
\end{align}

\subsection{Layout Generation}
\begin{figure}
    \centering
    \includegraphics[width=1\linewidth]{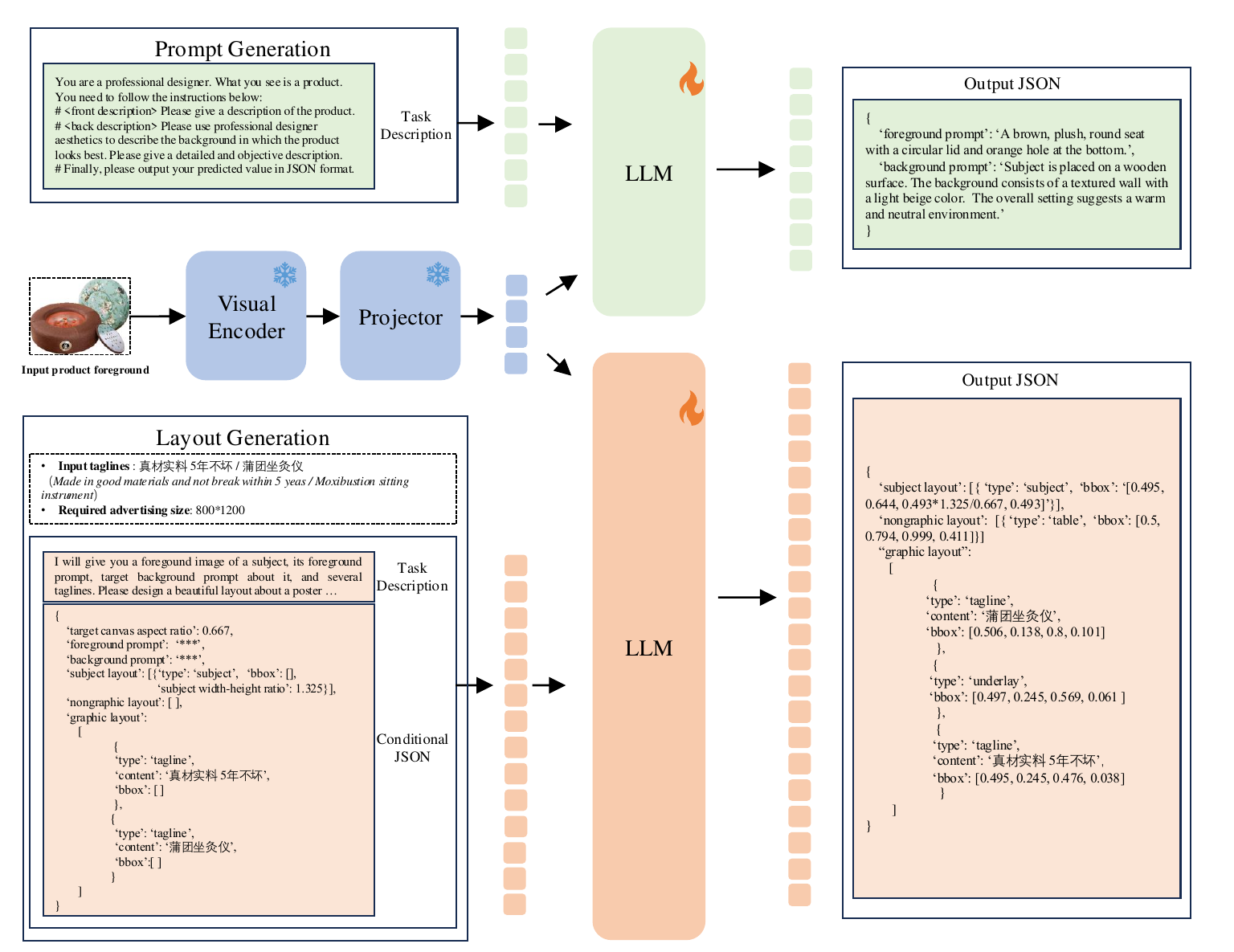}
    \vspace{-0.4cm}
    \caption{The framework of prompt and layout generation.}
    \label{fig:layoutGen}
    \vspace{-0.6cm}
\end{figure}

In this stage, we introduce the Jointly Predict Graphic and Nongraphic Layout (JPGNL) method, which optimizes the arrangement and improves the tagline readability and image appeal. It predicts the layout of graphic elements, the product, and other nongraphic elements based on the foreground image, prompt, taglines, and target size. Similar to prompt generation, we use XCP2~\cite{dong2024internlm} for layout generation. The method is shown in the lower part of Figure~\ref{fig:layoutGen}.

\textbf{Input data format.} For LLM input, we describe the layout generation task and organize conditions in JSON format. The image input is the same as in prompt generation. A target canvas aspect ratio is provided to support multi-scale design. The aspect ratio of the product foreground is given to keep its shape. Foreground/background prompts are included for image layout generation, and taglines are listed with empty locations for the model to fill in. If a logo is needed, it will be included with its aspect ratio.

We also take product characteristics into account. In e-commerce, some products need to be fully visible, while partial occlusion is acceptable for others, like clothing. To address this, we use a Class-Conditioned Layout Prediction (CCLP) strategy that indicates the product class and whether occlusion is allowed. For the not-allowing occlusion set (No Occ Set), we describe ``The class of subject is [V1]. The bounding boxes of taglines should never occlude the subject''. For the allowing occlusion set (Allow Occ Set), the description is ``The class of subject is [V2]. The bounding boxes of taglines are allowed to occlude the subject''.

\begin{figure}
    \centering
    \includegraphics[width=0.93\linewidth]{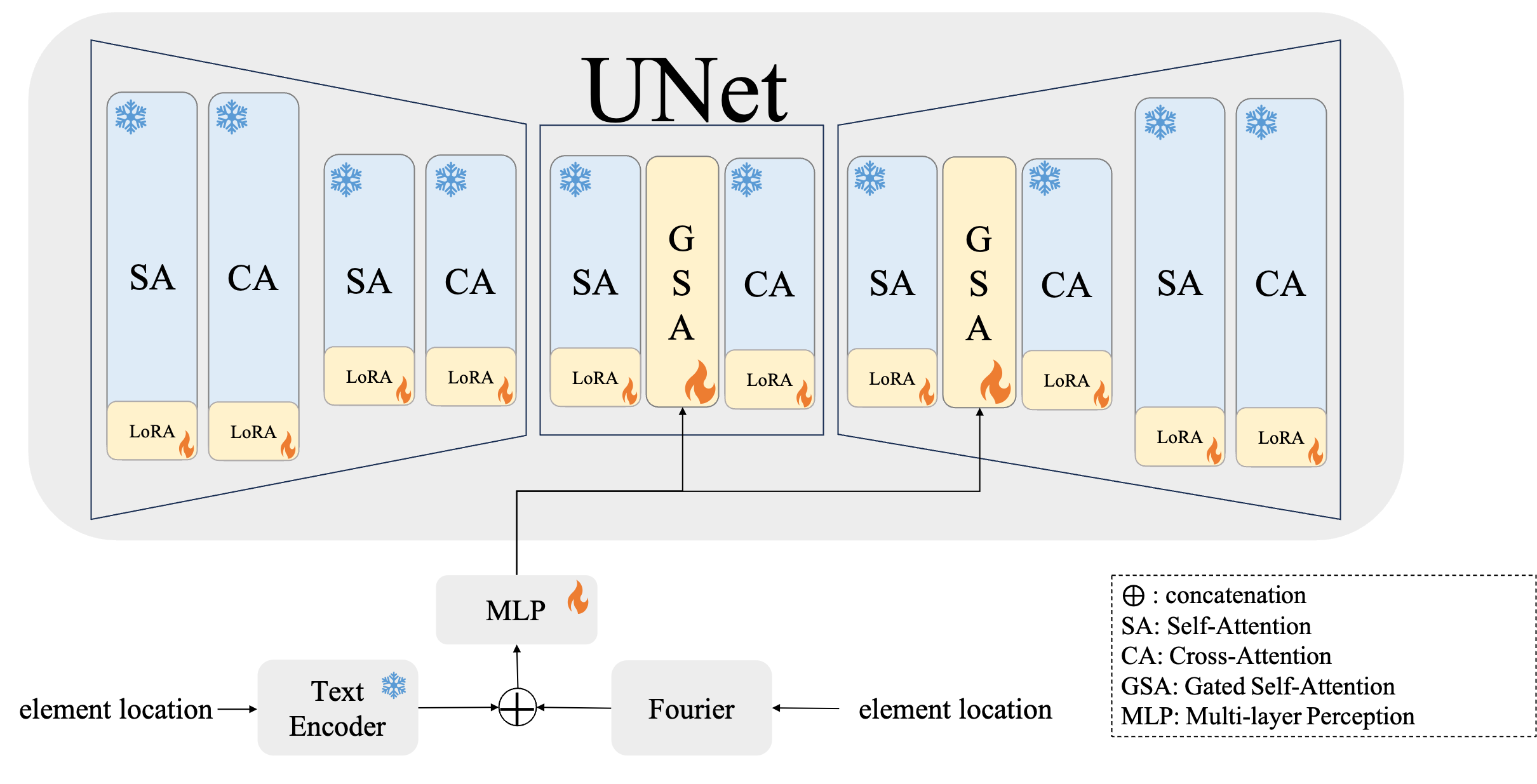}
    \vspace{-0.1cm}
    \caption{Layout control in the image generation model.}
    \label{fig:l2i_method}
    \vspace{-0.45cm}
\end{figure}

\textbf{Output data format.} The output layout is structured in JSON format with three parts: subject layout, other nongraphic layout, and graphic layout. To maintain the foreground shape, we propose a Ratio-Keeping Bbox Representation (RKBR): [\textit{x}, \textit{y}, \textit{h}*\textit{r}1/\textit{r}2, \textit{h}]. (\textit{x,y}) is the normalized center point. \textit{w} and \textit{h} are the normalized width and height. \textit{r}1 and \textit{r}2 are the actual aspect ratio of the foreground and canvas, respectively. Each value ranges from 0 to 1. The nongraphic layout contains the location of elements in the background prompt, each having a type~(instance name) and bounding box~(bbox). The graphic layout defines each element by type, content, and bounding box, covering the tagline, underlay, and logo. The content key is used only for tagline types to specify its semantic content. Bounding boxes are listed as [\textit{x}, \textit{y}, \textit{w}, \textit{h}].  The model is expected to supplement underlay adaptively. If logos are required, it is represented in the same way as the foreground for ratio keeping.

\textbf{Training scheme.}
Like the prompt generation module, we fine-tune only the LLM branch. The output JSON predicts layouts in a raster scan order, following ascending top-left coordinates. We also shuffle taglines in the input JSON for data augmentation to prevent the model from memorizing the sequence.

\subsection{Background Image Generation}
In this process, we train an SDXL-based~\cite{podell2023sdxl} image generation model controlled by product foreground, prompt, and layout.

\textbf{Layout control.} Suppose an image has $N$ elements, represented as $\{(n_i, l_i)|i=0,1,...,N-1\}$. We map these into $N$ embeddings as layout input. Here, $n_i$ is the element name, and $l_i$ is its location. We use the text encoder in SDXL to map $n_i$ into a global semantic embedding $\textbf{e}_{n,i}$. $l_i$ is transferred into normalized top-left and bottom-right coordinates, then fed into a Fourier mapping layer\cite{tancik2020fourier}, deriving the location embedding $\textbf{e}_{l,i}$. Next, $\textbf{e}_{n,i}$ and $\textbf{e}_{l,i}$ are concatenated and fed into a trainable MLP layer to represent the layout information of the $i$-th element, denoted as $\textbf{e}_{i}$. $\textbf{e} = \{\textbf{e}_i|i=0,1,...,N-1\}$ is the layout embeddings. A gated self-attention (GSA) layer~\cite{li2023gligen} is inserted between the self- and cross-attention layers in the original UNet to inject layout embeddings. If there are $M$ visual embeddings (${\textbf{V}}$) after the self-attention layer, the GSA output is
\begin{quote}
    $GSA(\textbf{V},\textbf{E}) = \textbf{V} + tanh(\gamma)* SA(cat(\textbf{V},\textbf{E}))[:M]$,
\end{quote}
where $\gamma$ is a learnable parameter, $cat$ means concatenation, and $SA$ means self-attention.

Since layout relates to semantic-level information and not high-frequency details, GSA layers are only added in the deep blocks of UNet. We refer to this strategy as Deep Layer Control (DLC). Specifically, the layout control is applied in the middle block and the lowest resolution of up blocks. Additionally, we use a LoRA Adaptation Training (LAT) strategy, adding LoRA~\cite{hu2021lora} layers to UNet to bridge the gap between the base model and added GSA layers. These further accelerate convergence and increase generation quality. Figure~\ref{fig:l2i_method} describes how layout control is achieved.

\textbf{Training stages.}
To build the product reservation layout-to-image model, training involves two steps. First, we modify the UNet architecture in SDXL by adding the above GSA layers and train it to derive a layout-to-image model. We use a multi-scale training strategy for faster convergency, first training on 512x512 resolution images and then fine-tuning on 1024x1024 resolution images. Second, we equip the trained model with an inpainting ControlNet and further fine-tune it on 1024x1024 resolution images. This step incorporates the product foreground, achieving full control over the prompt, layout, and product foreground.


\subsection{Graphics Rendering}
Since there is no effective method to predict the visual attributes of graphic elements, and the graphics rendering module is not the focus of this paper, we design a rendering strategy. We sort taglines by area to select the font and color from a limited set~\cite{liu2024glyph}. The choice depends on the similarity to the foreground color and contrast with the background. Taglines grouped by size and position use the same font and color. The underlay color is picked for contrast with the tagline and similarity to the background, while the shape is selected from a predefined SVG library based on size and proportions, with slight modifications to fit.



\section{Experiments}
In this section, we compare our framework with previous models.

\subsection{Implementation Details}
We use XCP2~\cite{dong2024internlm} as the backbone for prompt and layout generation, fine-tuning the LLM branch on the PITA dataset for one epoch respectively. The model is trained with the AdamW optimizer at a 1e-5 learning rate, using a total batch size of 64. Training takes 14 hours on 16 NVIDIA H20 GPUs. For inference, we apply top-p sampling with p set to 0.9 and a sampling temperature of 0.6.

For background image generation, we use the SDXL model as the backbone. In step 1, we train new layers using 6.8 million aesthetic images~(from the LAION~\cite{DBLP:conf/nips/SchuhmannBVGWCC22} and PIL dataset) with layout annotations at a learning rate of 5e-5 for 20k steps on 512x512 resolution, taking 65 hours on 16 NVIDIA H20 GPUs~(batch size 64). We then fine-tune them with a learning rate of 2e-5 for 20k steps at 1024x1024 resolution over 86 hours~(batch size 32). In step 2, we introduce an inpainting ControlNet pre-trained on advertising images~\cite{ecomxl_inpaint}. We fine-tune the layout-related layers with a learning rate of 2e-5 for 1k steps using 12k training data from the PIL dataset, which takes 5 hours with 16 NVIDIA H20 GPUs~(batch size 16).

\subsection{Evaluation Metrics}
We use the following metrics to assess the pipeline and each module.

\textbf{Overall pipeline.} We assess the pipeline from two perspectives: visual quality and layout. For visual quality, we use the Fréchet Inception Distance (FID)~\cite{heusel2017gans} and an aesthetic score, with the latter based on rankings by advertising experts. A lower average rank indicates better results. For layout evaluation, we use metrics from previous work~\cite{zhou2022composition,seol2024posterllama}, as detailed in the layout generation module.


\textbf{Prompt generation.} We evaluate prompt generation quality based on fore-background matching rate~(FBM rate) and e-commerce domain score~(ED score). FBM Rate measures how well a product foreground fits with the background prompt. We classify this matching into ``reasonable'' or ``unreasonable'' through human annotation, and calculate the proportion of reasonable cases. ED Score assesses how well the model-generated prompts reflect e-commerce characteristics. We calculate the FID between the CLIP features of these prompts and the ground truth prompts in the test set.

\begin{figure*}[h]
\centering
\includegraphics[width=0.9\textwidth]{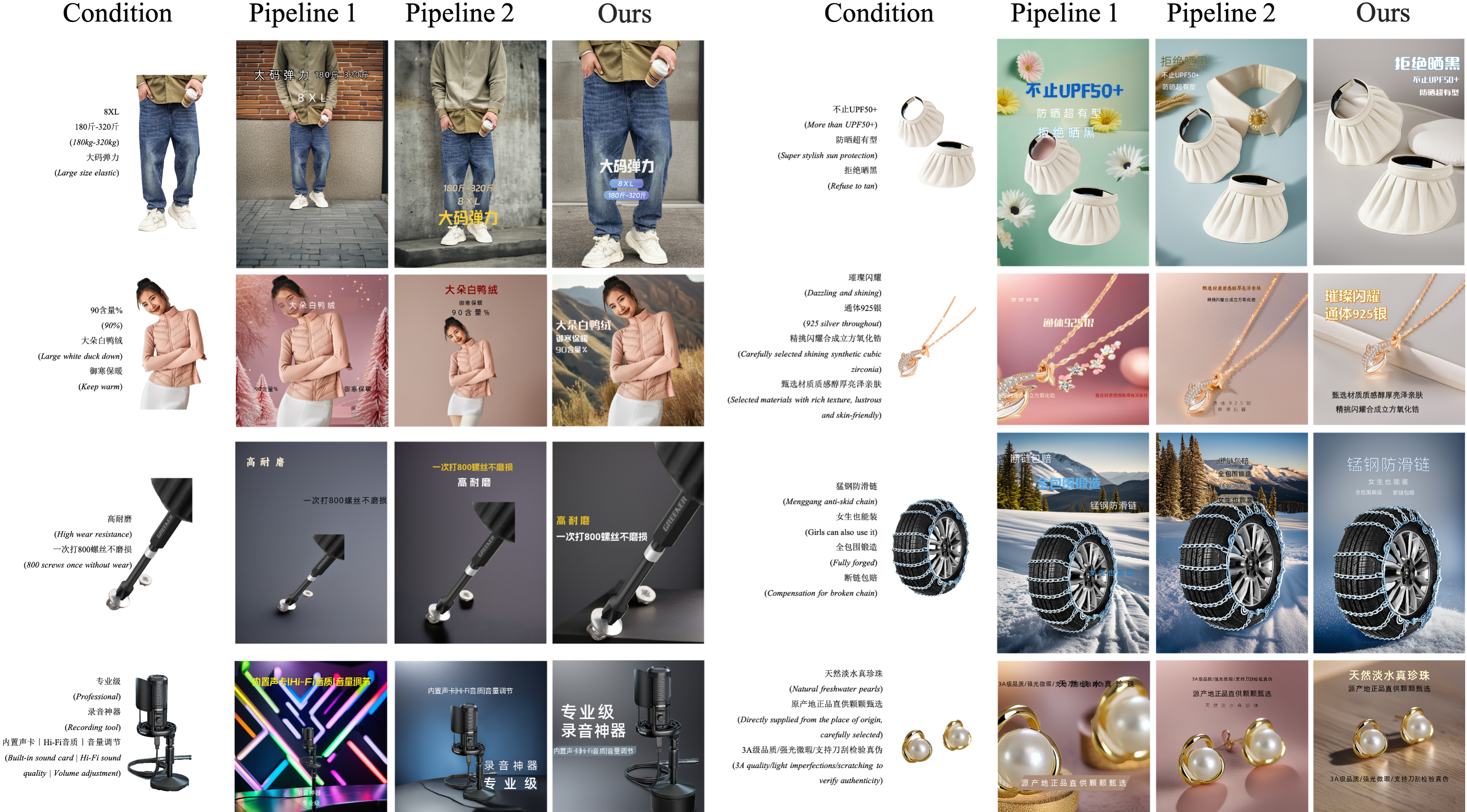}
\vspace{-0.2cm}
\caption{Visualization of advertising images designed by different methods.}
\vspace{-0.2cm}
\label{fig:pipeline_comp}
\end{figure*}

\textbf{Layout generation.} Following previous work~\cite{zhou2022composition,seol2024posterllama}, we use graphic/content metrics to evaluate layouts. Graphic metrics focus on relationships between graphic elements, including validity \(Val\), alignment \(Ali \), overlap \(Ove\), and underlay \(Und_l, Und_s\). 
Content metrics assess the harmony of layout with the image, including utility \(Uti\), occlusion \(Occ\), and unreadability \(Rea\). 
For methods that take taglines as input, we report the tagline match rate (TMR), to see if the number of generated tagline bboxes matches the input taglines.

\textbf{Background image generation.}
We evaluate the quality of generated images with FID, CLIP-T~\cite{ruiz2023dreambooth}, CLIP-I~\cite{gal2022image}. To verify layout control, we use Grounding-DINO~\cite{liu2024grounding} to detect instances and calculate the mIOU between detected bboxes and ground-truth bboxes. In addition, Grounding-DINO AP scores are also reported.

\subsection{Comparison}

\textbf{Compare with automatic pipelines for advertising image design. }
Since there are no public methods for advertising image design conditioned on the product foregrounds, taglines, and target sizes, we design two pipelines for comparison. \textbf{Pipeline 1} use GPT-4o~\cite{gpt4o} for prompt and layout generation. It fine-tunes an SDXL-based inpainting model on our PIL dataset. The main difference from our approach is the lack of layout control. During inference, only the product foreground is controlled by the layout output of GPT-4o. \textbf{Pipeline 2} first generate a background prompt using GPT-4o. The image generation model is the same as in Pipeline 1, but during inference, the foreground is placed according to general aesthetic rules. Next, the image and taglines are processed by the SOTA layout generation model PosterLlama to generate a graphic layout. Last, the results of these two pipelines are processed through the same graphics rendering module as our method to produce the final image. Table~\ref{tab:comp_pipelines} 
shows the quantitative comparison.
T-Stars-Poster outperforms in most metrics, except for slightly higher occlusion compared to Pipeline 1, which places the product centrally with a small bounding box. However, Pipeline 1 uses GPT-4o for layout, lacking product-centric design knowledge, resulting in lower quality. T-Stars-Poster uses trained experts for different tasks in ad design, creating more visually appealing results (see Figure~\ref{fig:pipeline_comp}). Pipeline 2 fixes the product location, limiting adaptability to product shapes, whereas T-Stars-Poster adjusts layouts based on product shape for better visuals. Additionally, the fixed background of Pipeline 2 can crowd tagline boxes, reducing readability, while our model generates more flexible and readable layouts.

\begin{table}[h]
    \vspace{-0.1cm}
    \centering
    \caption{Quantitative comparison between automatic pipelines for product-centric advertising image design.}
    \vspace{-0.2cm}
    \label{tab:comp_pipelines}
    \scalebox{0.63}{
    \begin{tabular}{cccccccccc}
    \toprule
    & \multicolumn{2}{c}{Overall Visual Quality} & \multicolumn{7}{c}{Layout Quality} \\
    \cmidrule(lr){2-3} \cmidrule(lr){4-10}
    Method  & Fid$\downarrow$ & Aesthetic$\downarrow$ & Ove$\downarrow$ & Ali$\downarrow$ & $Und_l\uparrow$ & $Und_s\uparrow$ & Uti$\uparrow$ & Occ$\downarrow$ & Rea$\downarrow$ \\
    \midrule
    Pipeline 1  & 56.545 & 2.781 & 0.0021 & 0.002 & 0.8333 & 0.7872 & 0.0891 & \textbf{0.0937} & 0.234 \\
    Pipeline 2  & 43.221 & 1.814 & 0.0016 & 0.0028 & 0.9994 & 0.9930 & 0.0984 & 0.1158 & 0.1864 \\
    T-Stars-Poster & \textbf{37.524} & \textbf{1.405} & \textbf{0.0013} & \textbf{0.0017} & \textbf{0.9999} & \textbf{0.9973} & \textbf{0.1367} & 0.0955 & \textbf{0.1815} \\
    \bottomrule
    \vspace{-0.5cm}
    \end{tabular} 
    }
\end{table}


\textbf{Compare with prompt generation methods.}
For prompt comparison, we use GPT-4o and an untrained version of our model~(XCP2) as baselines. Both utilize in-context learning (ICL) for optimal performance. The FBM Rate is determined by votes from three annotators for each case. As shown in Table~\ref{prompt_table}, our method outperforms the baselines, which means that our model effectively aligns with the e-commerce prompt dataset distribution after training.

\begin{table}[h]
    \centering
    \vspace{-0.2cm}
    \caption{Evaluation of prompt generation methods.}
    \vspace{-0.2cm}
    \scalebox{0.85}{
    \begin{tabular}{ccc}
        \toprule
        Method  & FBM Rate $\uparrow$ & ED Score $\downarrow$ \\
        \midrule
        GPT-4o & 97.9\% & 22.49 \\
        XCP2 & 94.8\% & 13.04 \\
        Finetuned XCP2  & \textbf{98.5\%} & \textbf{7.17} \\
        \bottomrule
        \vspace{-0.6cm}
    \end{tabular}
    }
    \label{prompt_table}
\end{table}

\begin{figure*}[htbp]
    \centering
    \includegraphics[width=0.9\linewidth]{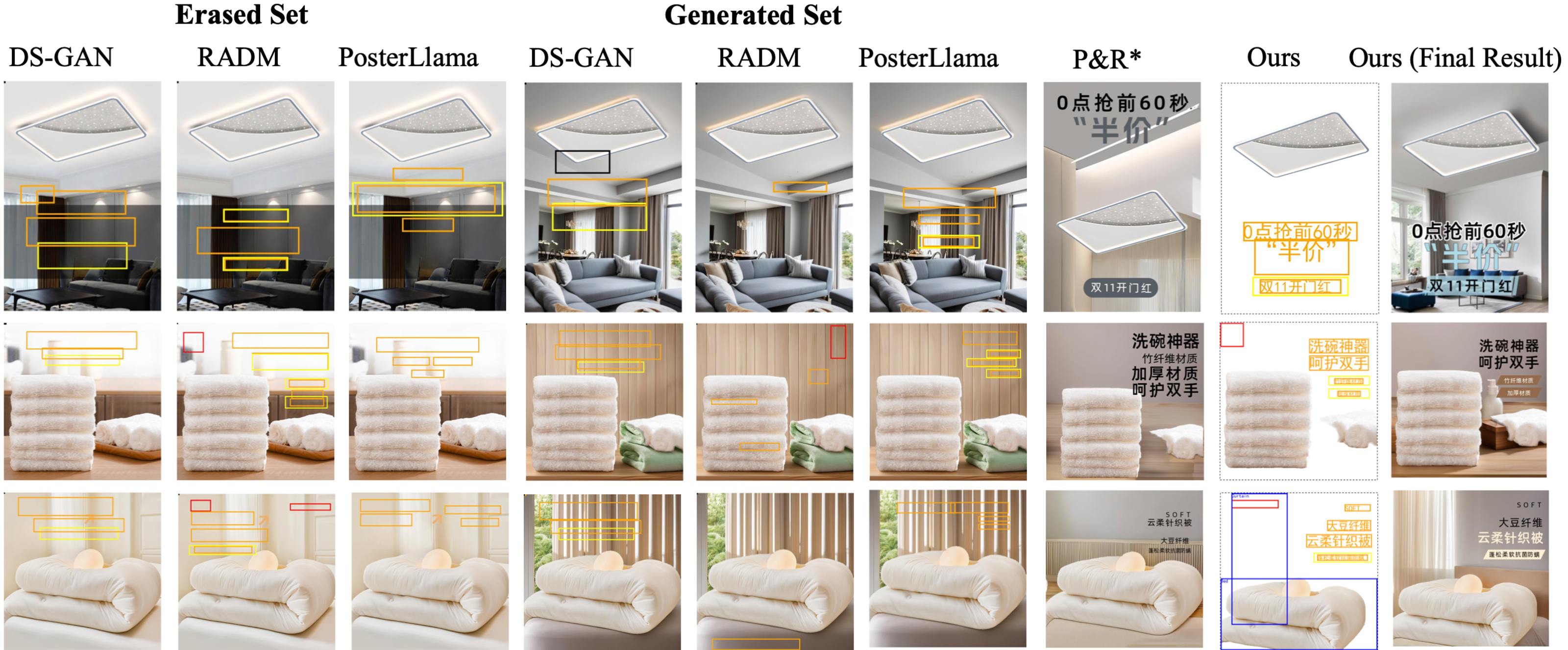}
    \vspace{-0.2cm}
    \caption{Qualitative comparison of layout generation methods. Orange: tagline, Yellow: underlay, Red: logo, Black: Invalid.}
    \vspace{-0.2cm}
    \label{fig:layout_generation_sota}
\end{figure*}

\textbf{Compare with layout generation methods.}
We compare our methods with background-based methods~(DS-GAN~\cite{hsu2023posterlayout}, RADM~\cite{li2023relation}, and PosterLlama~\cite{seol2024posterllama}) and the foreground-based method, P\&R~\cite{li2023planning}. For background-based methods, we test on two sets for fair comparison: the erased set, with graphic elements erased by LaMa~\cite{suvorov2022resolution}, which may provide hints to put graphic elements on erased areas; and the generated set, which contains re-generated images with their ground-truth prompts~(image captions), aligning with real applications. Since P\&R has not released the codes, we re-implemented its idea. Ground-truth prompts are also given to P\&R and our method.
The results are shown in Table~\ref{tab:layout_generation} and Figure~\ref{fig:layout_generation_sota}. Background-based methods struggle with product occlusion, overlap, and alignment of graphic elements, likely due to fixed background content. RADM and DS-GAN also have issues with tagline boxes that cannot properly fit the content. P\&R improves in these areas but sometimes generates unreasonable images due to a lack of harmony between product, background prompt, and graphic layout. T-Stars-Poster considers product shape, background prompt, and tagline content, optimizing the entire layout for graphic and non-graphic elements, ultimately achieving better overall results.


\begin{table}[h]
    \centering
    \vspace{-0.1cm}
    \Large
    \setlength{\tabcolsep}{1.4pt}
    \caption{Quantitative performance comparison of layout generation methods. The bold number represents the best result in each column except for the erased set. BG/FG-based are the abbreviations for Background/Foreground-based methods.}
    \vspace{-0.2cm}
    \label{tab:layout_generation}
    \scalebox{0.56}{
    \begin{tabular}{ccccccccccc}
    \toprule
    \multicolumn{2}{c}{Method} & Val$\uparrow$ & Ove$\downarrow$ & Ali$\downarrow$ & $Und_l\uparrow$ & $Und_s\uparrow$ & Uti$\uparrow$ & Occ$\downarrow$ & Rea$\downarrow$ & TMR$\uparrow$ \\
    \midrule
    \multirow{3}*{\makecell{BG-based\\(Erased)}} & DS-GAN\cite{hsu2023posterlayout} & 0.9585 & 0.0270 & 0.0058 & 0.3910 & 0.0744 & 0.1816 & 0.1063 & 0.1826 & - \\
    ~ & RADM\cite{li2023relation} & 0.999 & 0.0411 & 0.0017 & 0.9852 & 0.6934 & 0.1484 & 0.0814 & 0.1722 & 0.82 \\
    ~ & PosterLlama\cite{seol2024posterllama} & 0.9984 & 0.002 & 0.0026 & 0.9899 & 0.9870 & 0.1226 & 0.0976 & 0.1820 &0.998 \\
    \hline
    \multirow{3}*{\makecell{BG-based\\(Generated)}} & DS-GAN\cite{hsu2023posterlayout} & 0.9621 & 0.0284 & 0.0080 & 0.3350 & 0.0528 & \textbf{0.1701} & 0.1202 & 0.2328 & - \\
    ~ & RADM\cite{li2023relation} & 0.9753 & 0.0484 & 0.0201 & 0.7997 & 0.2828 & 0.0657 & 0.2527 & 0.2558 & 0.289 \\
    ~ & PosterLlama\cite{seol2024posterllama} & \textbf{1.0} & 0.0015 & 0.0022 & \textbf{0.9990} & \textbf{0.9965} & 0.1090 & 0.1208 & 0.231 & 0.996 \\
    \hline
    \multirow{3}*{FG-based} & P\&R$^{*}$\cite{li2023planning} & \textbf{1.0} & \textbf{0.0012} & 0.0019 & 0.9966 & 0.9929 & 0.1367 & 0.1000 & 0.2008 & \textbf{1.0} \\
    ~ & Ours & \textbf{1.0} & \textbf{0.0012} & \textbf{0.0017} & 0.9976 & 0.9956 & 0.1364 & \textbf{0.0973} & \textbf{0.1968} & \textbf{1.0} \\
    \bottomrule 
    \vspace{-0.5cm}
    \end{tabular}
    }
\end{table}


\textbf{Compare with layout-controlled inpainting models.}
We compare our layout-controlled background image generation model with others: the training-based method GLIGEN~\cite{li2023gligen} and training-free methods like BoxDiff~\cite{xie2023boxdiff}, TFCLG~\cite{chen2024training}, and MultiDiffusion~\cite{bar2023multidiffusion}. Since these methods lack inpainting ability, we re-implement them with the SDXL-based inpainting ControlNet~\cite{ecomxl_inpaint}. The results are reported in Table~\ref{tab:layout2image}. 
Our trained model outperforms others in FID, mIoU, and AP, indicating better image quality and spatial control. 



\begin{table}[h]
    \centering
    \vspace{-0.15cm}
    \caption{Quantitative comparison of layout-controlled inpainting models on PIL test set.}
    \vspace{-0.3cm}
    \label{tab:layout2image}
    \scalebox{0.78}{
    \begin{tabular}{cccccc}
    \toprule
    Method & FID$\downarrow$ & CLIP-T$\uparrow$ & CLIP-I$\uparrow$ & mIoU$\uparrow$ & AP/AP50/AP75$\uparrow$ \\
    \midrule
    BoxDiff\cite{xie2023boxdiff} & 29.383 & 0.315 & 0.896 & 0.555 & 0.037/0.064/0.033 \\
    TFCLG\cite{chen2024training} & 29.498 & 0.315 & 0.895 & 0.543 & 0.033/0.056/0.029 \\
    MultiDiffusion\cite{bar2023multidiffusion} &42.359 & 0.311 & 0.841 & 0.5617 & 0.029/0.052/0.025 \\
    GLIGEN\cite{li2023gligen}  & 27.237 & 0.312 & 0.898 & 0.561 & 0.047/0.073/0.044 \\
    Ours  & \textbf{25.917} & \textbf{0.313} & \textbf{0.906} & \textbf{0.705} & \textbf{0.079/0.127/0.078} \\
    \bottomrule
     \vspace{-0.7cm}
    \end{tabular}
    }
\end{table}

\subsection{User-Centric Study}
To thoroughly evaluate the visual effects and attractiveness of the generated results, we randomly select 200 test samples and ask 7 Taobao users to compare images in pairs, evaluating background aesthetics, layout harmony, text readability, and overall visual appeal. As shown in Table~\ref{tab:user-centric study}, our method significantly outperforms the comparison methods in all aspects.

\begin{table}[h]
    \centering
    \vspace{-0.15cm}
    \caption{User-centric study. The values in the table represent the winning rate of our method compared to other methods.}
    \vspace{-0.3cm}
    \label{tab:user-centric study}
    \scalebox{0.69}{
    \begin{tabular}{ccccc}
    \toprule
    Comparison Method & Bg Aesthetic & Layout Harmony & Text Readability & Visual Appeal \\
    \midrule
    Pipeline 1 & 84\% & 93\% & 94\% & 91\% \\
    Pipeline 2 & 77\% & 85\% & 89\% & 74\% \\
    \bottomrule
     \vspace{-0.6cm}
    \end{tabular}
    }
\end{table}

\subsection{Online Result}
To evaluate the online performance of T-Stars-Poster, we conduct A/B tests in two advertising recommendation scenarios on Taobao. We randomly select 5,000 products and collect their foreground images and taglines from advertisers.
We generate images for each product using both T-Stars-Poster and Pipeline 2 (the top-performing baseline as shown in Table~\ref{tab:comp_pipelines}). The A/B tests use 5\% of the main traffic, affecting only the experimental products when this traffic is directed to them. After gathering data over one month, results show that T-Stars-Poster achieved a 3.02\% and 3.03\% increase in click-through rate (CTR) in the two recommendation scenarios, respectively. This indicates that T-Stars-Poster produces more visually appealing advertising images, resulting in improved recommendation outcomes.

\section{Limitations and Discussion}
Through observation, we have found that the results may be unsatisfactory in some cases. 1) Difficulty in adjusting product angle and lighting. As shown in Figure~\ref{fig:pipeline_comp}, inpainting models cannot adjust the product angle and lighting due to the model structure and training data, possibly causing disharmony in lighting and composition. While subject-driven methods~\cite{DBLP:journals/corr/abs-2411-15098} may help, they currently cannot fully preserve product features. 2) Limited flexibility with multiple foregrounds. As Figure~\ref{fig:pipeline_comp} shows, multiple foregrounds are treated as one entity. This limits adjustments when products are spaced apart or angled very differently. Future improvements could involve treating separate subjects individually
and introducing position augmentations during training. 
3) Poor layout with many taglines. Over five taglines increase overlap and affect readability due to limited training data (4.3\%) and canvas space. More data with multiple taglines might help resolve this. 

\section{Conclusion}
This paper introduces T-Stars-Poster, a product-centric framework for creating advertising images using product foregrounds, taglines, and target sizes. It comprises four stages: prompt generation, layout generation, background image generation, and graphics rendering. First, it generates prompts according to the product foregrounds. Then, it predicts how graphic and nongraphic elements should be placed according to the prompt, product foreground, and taglines, creating a harmonious overall layout. Next, a layout-controlled inpainting model is utilized for background image generation. A graphics rendering module is applied to get the final images. We train separate experts to conduct these sub-tasks. Two datasets are created for the convenience of training and testing. Test results and online A/B tests show that T-Stars-Poster produces more visually pleasing and attractive advertising images.

\section*{Acknowledgements}
This work is supported by Alibaba Research Intern Program.

\newpage
\bibliographystyle{ACM-Reference-Format}
\bibliography{sample-base}



\newpage
\appendix

\section{Dataset Examples}
\begin{figure}[h!]
    \centering
    \includegraphics[width=0.9\linewidth]{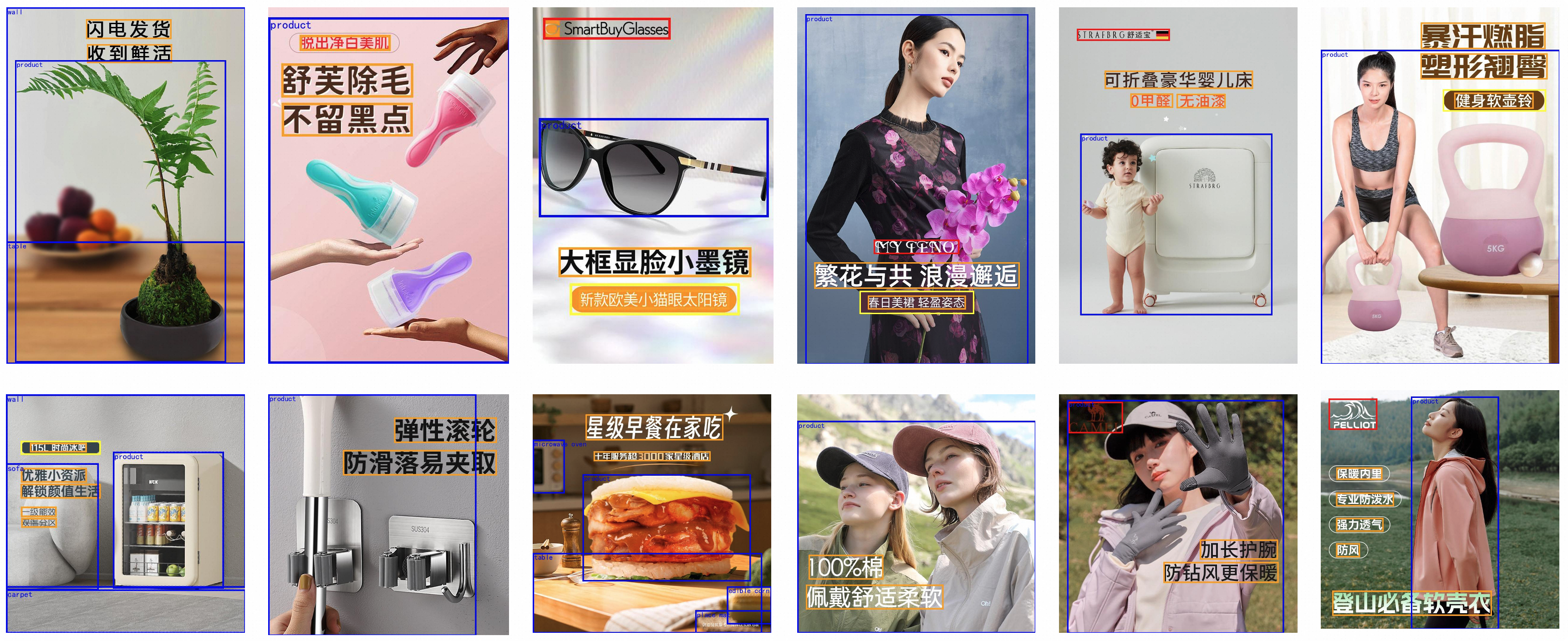}
    \caption{Examples from the PITA dataset, with tagline, underlay, and logo highlighted in orange, yellow, and red rectangles, respectively.}
    \label{fig:dataset}
    \vspace{-0.35cm}
\end{figure}

\begin{figure}[h!]
    \centering
    \includegraphics[width=1\linewidth]{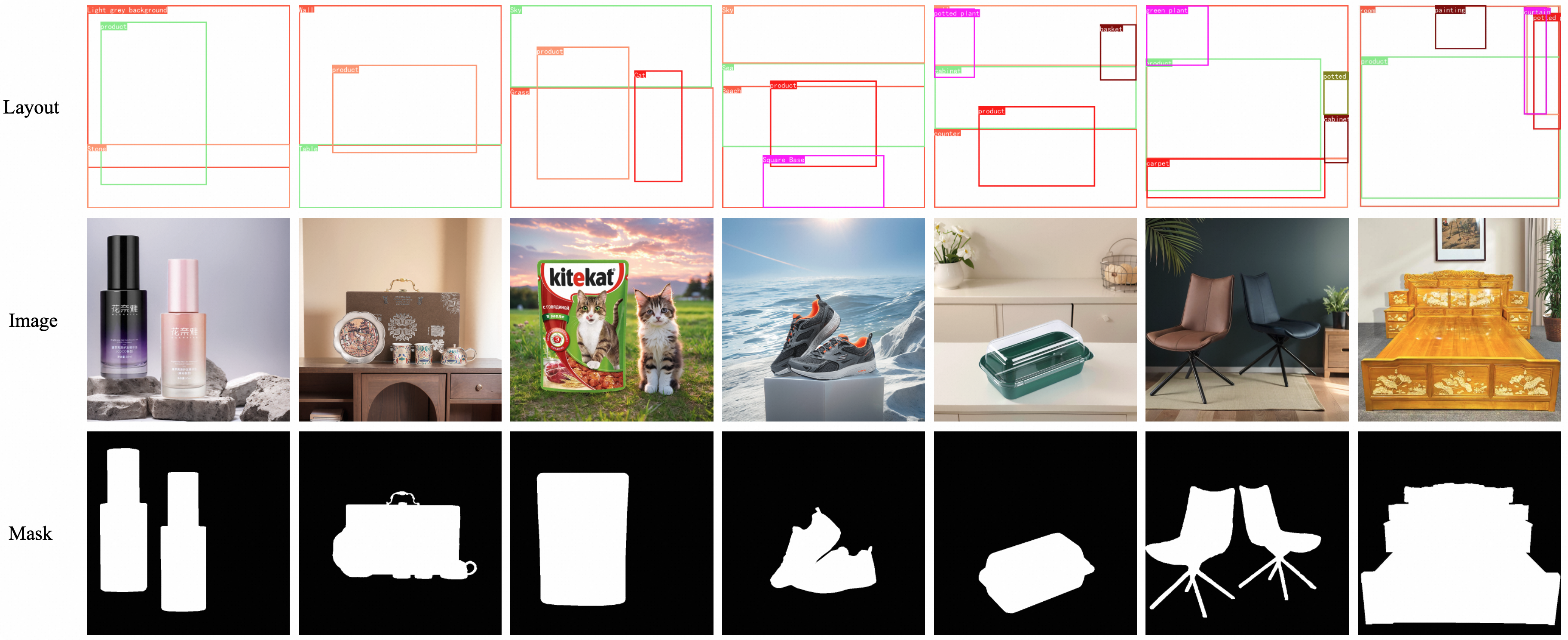}
    \caption{Examples in PIL dataset.}
    \label{fig:dataset2}
\end{figure}

In this section, we provide examples from the PITA and PIL datasets, illustrated in Figures~\ref{fig:dataset} and~\ref{fig:dataset2}, respectively. The PITA dataset includes advertising images with marketing taglines. In addition to depicting the layout of both graphic and nongraphic elements as shown in the figures, we also label the image caption~(prompt), tagline content, and product foreground masks to train the prompt and layout generation models. Conversely, the PIL dataset contains product images without taglines. We annotate the layout of nongraphic elements and product foreground masks for training the image generation model.

\section{Visualization of Prompt Distribution}
\begin{figure}[h!]
\centering
\includegraphics[width=0.34\textwidth]{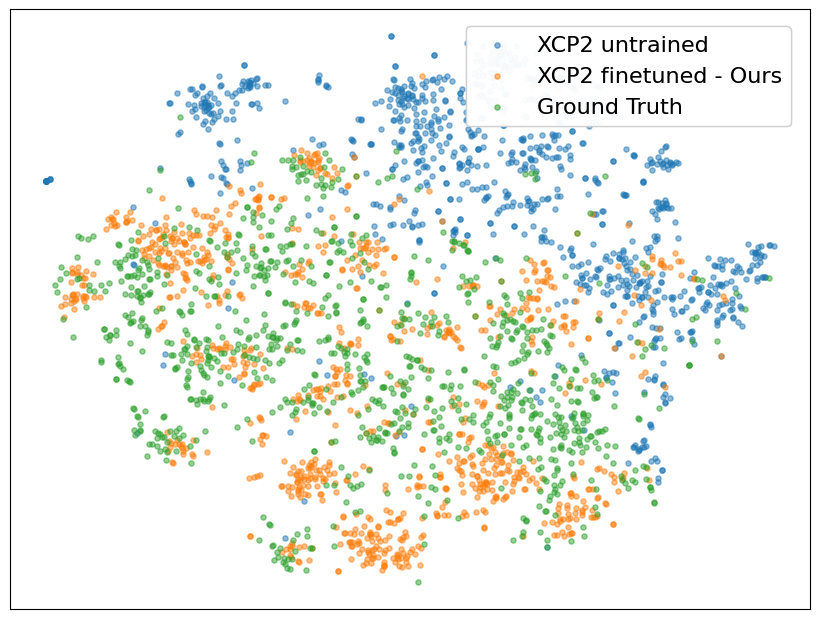}
\caption{T-sne visualization of generated prompts.}
\label{fig:tsne}
\end{figure}

We use t-SNE for dimensionality reduction to visualize the feature distribution of prompts generated by our model before and after training (see Figure~\ref{fig:tsne}). This shows that our model effectively aligns with the e-commerce prompt dataset distribution after training.

\section{Ablation Studies}
To verify the effectiveness of our design, we conduct ablation studies on the methods and strategies used in each module.

\subsection{Effect of JPGNL in Layout Generation} 
\begin{figure}
\centering
\includegraphics[width=0.49\textwidth]{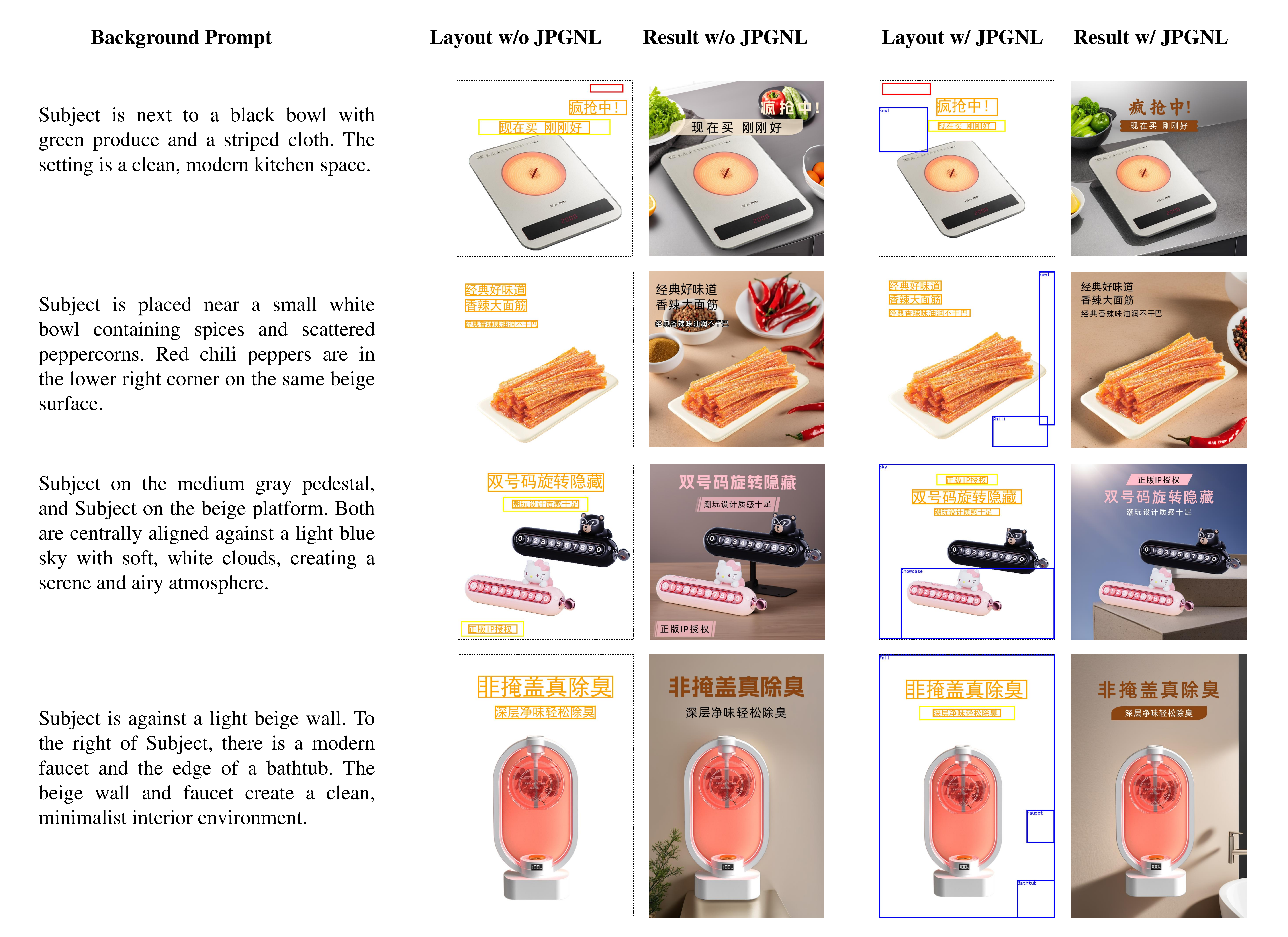}
\caption{An illustration of the effect of JPGNL. Predicting image layout makes better visual effect.}
\label{fig:compare_without_imgele}
\end{figure}

Our method predicts the overall layout of graphic and nongraphic elements. Here, we examine the need to predict the layout of nongraphic elements according to the background prompt. As shown in Figure~\ref{fig:compare_without_imgele}, without specifying the position of nongraphic elements, taglines placed on complex areas can lead to a messy appearance. By predicting the overall layout, we improve the tagline readability of advertising images. Additionally, with nongraphic layout predictions, generated images align more closely with background prompts. Without JPGNL, images might miss some objects from the prompts. Besides, in Table~\ref{tab:predict_imglayout}, we quantitatively analyze the impact of predicting nongraphic layouts on graphic layouts. Predicting nongraphic layouts (w/ JPGNL) can improve metrics related to product occlusion and tagline readability. However, since it adds complexity to the task, it negatively affects graphic metrics such as element overlap and alignment.

\begin{table}[h!]
    \centering
    \caption{Quantitative ablation study on JPGNL.}
    \label{tab:predict_imglayout}
    \scalebox{0.7}{
    \begin{tabular}{ccccccccc}
    \toprule
    Method & Val$\uparrow$ & Ove$\downarrow$ & Ali$\downarrow$ & $Und_l\uparrow$ & $Und_s\uparrow$ & Uti$\uparrow$ & Occ$\downarrow$ & Rea$\downarrow$  \\
    \midrule
    w/o JPGNL  & 1.0 & \textbf{0.0009} & \textbf{0.0013} & \textbf{0.9995} & 0.9891 & \textbf{0.1377} & 0.1001 & 0.2003 \\
    w/ JPGNL & 1.0 & 0.0012 & 0.0017 & 0.9976 & \textbf{0.9956} & 0.1364 & \textbf{0.0973} & \textbf{0.1968} \\
    \bottomrule
    \end{tabular} 
    }
\end{table}

\subsection{Effect of RKBR and CCLP in Layout Generation} We validate the effectiveness of the proposed input and output format for layout generation, including RKBR and CCLP. As shown in Table~\ref{tab:ablation_data_format}, without RKBR, the model may predict a product size which does not match the original aspect ratio about 5\% of the time. Note that we regard a difference of less than 1.5\% as correctness. Adding CCLP improves the Uti and Occ values on No Occ Set, indicating that CCLP helps the model to distinguish occlusion-allowing sets and others by providing explicit class conditions.

\setlength{\tabcolsep}{2pt} 
\begin{table}[h!]
    \centering
    \caption{Ablation study on input/output format construction of layout generation. \textit{O}, \textit{NO}, and \textit{AO} correspond to the overall set, the No Occ Set, and the Allow Occ Set, respectively. FRC represents Fg Ratio Correctness.}
    \label{tab:ablation_data_format}
    \scalebox{0.61}{
    \begin{tabular}{cccccccccccc}
    \toprule
    RKBR & CCLP & Val$\uparrow$ & Ove$\downarrow$ & Ali$\downarrow$ & $Und_l\uparrow$ & $Und_s\uparrow$ & Uti(O$\uparrow${} /NO$\uparrow$ / AO) & Occ(O$\downarrow${} /NO$\downarrow${} /AO) & Rea$\downarrow$ & FRC$\uparrow$\\
    \midrule
    \XSolidBrush & \XSolidBrush & \textbf{1.0} & 0.0012 & 0.0018 & 0.9982 & \textbf{0.9971} & 0.1353/0.1448/0.0876 & 0.0985/0.0230/0.4258 & 0.2005 & 0.955 \\
    \Checkmark & \XSolidBrush & \textbf{1.0} & 0.0012 & 0.0018 & 0.9991 & 0.9912 & 0.1360/0.1466/0.0816 & 0.1054/0.0209/0.4491 & 0.1994 & \textbf{1.0} \\
    \XSolidBrush & \Checkmark & \textbf{1.0} & \textbf{0.0008} & 0.0018 & \textbf{0.9989} & 0.9906 & \textbf{0.1366}/0.1492/0.0851 & 0.1002/0.0156/0.4282 & 0.1973 & 0.957 \\
    \Checkmark & \Checkmark & \textbf{1.0} & 0.0012 & \textbf{0.0017} & 0.9976 & 0.9956 & 0.1364/\textbf{0.1493}/0.0849 & \textbf{0.0973}/\textbf{0.0136}/0.4316 & \textbf{0.1968} & \textbf{1.0} \\
    \bottomrule
    \end{tabular} 
    }
\end{table}

\subsection{Effect of DLC and LAT in Layout-to-Image Model} We investigate the effect of deep layer control~(DLC) and Lora adaptation training~(LAT), with results shown in Table~\ref{tab:ablation_layout2image}. Adding layout control in deep UNet layers performs almost as well as controlling all layers. It reduces parameters and inference costs, which benefits the application. With LoRA adaptation, the model achieves lower FID and higher mIoU, enhancing image quality and spatial control. Parameters and inference costs can be further reduced.

\begin{table}[h!]
    \centering
    \caption{Quantitative performance comparison of layout-to-image methods on PIL dataset.}
    \label{tab:ablation_layout2image}
    \scalebox{0.75}{
    \begin{tabular}{ccccccc}
    \toprule
    DLC & LAT & FID$\downarrow$ & CLIP-T$\uparrow$ & CLIP-I$\uparrow$ & mIoU$\uparrow$ & AP/AP50/AP75$\uparrow$ \\
    \midrule
    \XSolidBrush & \XSolidBrush & 27.237 & 0.312 & 0.898 & 0.561 & 0.047/0.073/0.044 \\
    \Checkmark & \XSolidBrush & 27.308 & 0.312 & 0.898 & 0.558 & 0.045/0.075/0.043 \\
    \XSolidBrush & \Checkmark & 25.675 & \textbf{0.313} & \textbf{0.906} & 0.696 & 0.076/0.123/0.075 \\
    \Checkmark & \Checkmark & \textbf{25.917} & \textbf{0.313} & \textbf{0.906} & \textbf{0.705} & \textbf{0.079/0.127/0.078} \\
    \bottomrule
    \end{tabular}
    }
\end{table}

\end{document}